\documentclass[sigconf]{acmart}
\usepackage{multirow}
\usepackage{subfigure}
\usepackage{balance}
\AtBeginDocument{%
  \providecommand\BibTeX{{%
    \normalfont B\kern-0.5em{\scshape i\kern-0.25em b}\kern-0.8em\TeX}}}

\setcopyright{acmcopyright}
\copyrightyear{2021} 
\acmYear{2021} 
\setcopyright{acmcopyright}
\acmConference[SIGIR '21]{Proceedings of the 44th International ACM SIGIR Conference on Research and Development in Information Retrieval}{July 11--15, 2021}{Virtual Event, Canada}
\acmBooktitle{Proceedings of the 44th International ACM SIGIR Conference on Research and Development in Information Retrieval (SIGIR '21), July 11--15, 2021, Virtual Event, Canada}
\acmPrice{15.00}
\acmDOI{10.1145/3404835.3462881}
\acmISBN{978-1-4503-8037-9/21/07}

\begin{document}
\fancyhead{} 
\title{Conversational Fashion Image Retrieval via Multiturn Natural Language Feedback}

\author{Yifei Yuan}
\email{yfyuan@se.cuhk.edu.hk}
\affiliation{%
  \institution{The Chinese University of Hong Kong}
}



\author{Wai Lam}
\email{wlam@se.cuhk.edu.hk}
\affiliation{%
 \institution{The Chinese University of Hong Kong}
 }






\begin{abstract}
  We study the task of conversational fashion image retrieval via multiturn natural language feedback. Most previous studies are based on single-turn settings. Existing models on multiturn conversational fashion image retrieval have limitations, such as employing traditional models, and leading to ineffective performance. We propose a novel framework that can effectively handle conversational fashion image retrieval with  multiturn natural language feedback texts. One characteristic of the framework is that it searches for candidate images based on exploitation of the encoded reference image and feedback text information together with the conversation history. Furthermore, the image fashion attribute information is leveraged via a mutual attention strategy. Since there is no existing fashion dataset suitable for the multiturn setting of our task, we derive a large-scale multiturn fashion dataset via additional manual annotation efforts on an existing single-turn dataset. The experiments show that our proposed model significantly outperforms existing state-of-the-art methods.

\end{abstract}


\thanks{The work described in this paper is substantially supported by a grant
from the Research Grant Council of the Hong Kong Special Administrative
Region, China (Project Codes: 14200719).}


\maketitle

\section{Introduction}
\label{intro}
Fashion is one of the most glamorous industries of the modern society, making great contributions to the global economy. With the development of image processing and information retrieval techniques, recently some investigation has been conducted in this domain, including fashion design~~\cite{rostamzadeh2018fashion,ma2017towards}, fashion product recommendation~\cite{han2017learning,liu2012hi,hu2015collaborative,li2017mining}, and conversational fashion image retrieval~\cite{guo2018dialog,guo2019fashion,zhang2020reward}.
\begin{figure}[]
  \centering
  \includegraphics[width=\linewidth]{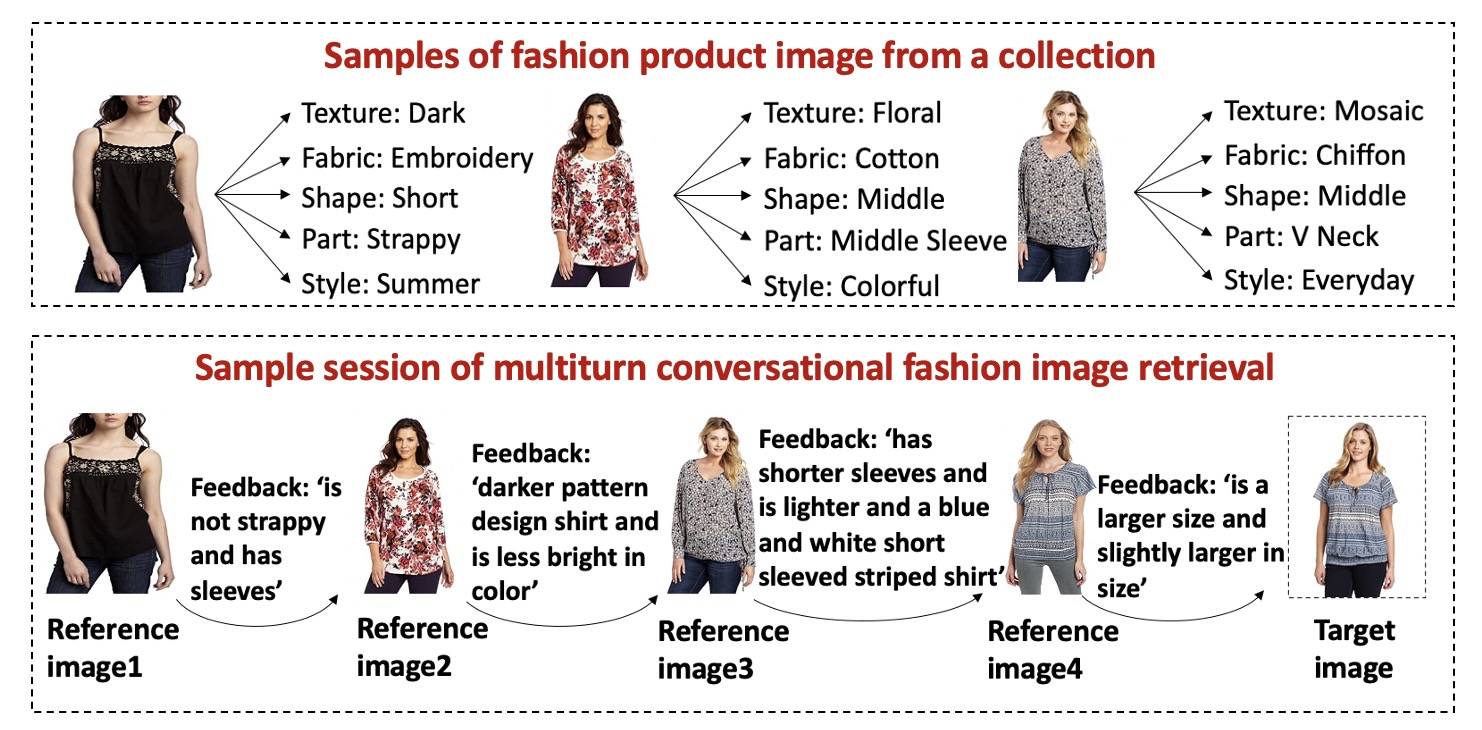}
  \caption{The top part is an example of three fashion product images. Each image has some fashion attributes. The bottom part is a sample session of multiturn conversational fashion image retrieval with natural language feedback.}
  \label{dataset}
\end{figure}
Building an interactive conversational fashion image retrieval system based on user feedback has drawn increasing research interests in the past years. One important task in interactive conversational fashion image retrieval is target image selection, which aims to find the best-matched fashion image, called target image, from a set of candidate fashion product images via interactions of intermediate retrieved images, called reference images, and user natural language feedback texts. Consider a collection of fashion product images as shown in the top part of Figure \ref{dataset}, each image is associated with some fashion attributes. Suppose that a user has an information need of a fashion product, after an initial interaction with the system, a fashion product image is retrieved and presented to the user. Based on this intermediate reference image, the user  typically wishes to refine the retrieval by providing natural language feedback texts, which describe the relative difference between the current retrieved reference image and the desired one. Such process is defined as a turn. If the user is not satisfied with the retrieved image, more turns are conducted until the desired product is retrieved. This multiturn process, consisting of several reference images, feedback texts as well as the final target image, is named as a session.

Studies on multimodal feature composition have shown great promise in single-turn conversational image retrieval. Treating the query as a composition of an image and a text, these methods tackle the task by combining the visual and language representations~\cite{noh2016image,718510,vo2019composing,anwaar2021compositional}.
These works, though having reasonable performance, are limited to single-turn feedback and cannot handle the multiturn conversational image retrieval task in our setting.

Recently some existing multiturn conversational image retrieval methods have been investigated,  however, their architecture and techniques are quite simple. ~\cite{kovashka2012whittlesearch} proposes a unique mode of feedback for image search, where users are allowed to give some property related binary feedback attempting to  match his/her mental model of the target image(s). Others try to retrieve the target image based on multiple relevance levels ~\cite{datta2008image} or relative attributes ~\cite{kovashka2012whittlesearch}. ~\cite{liao2018knowledge} proposes a knowledge-aware multimodal dialogue model which gives special consideration to the semantics and domain knowledge revealed in visual content.  ~\cite{guo2018dialog} first introduces a deep learning based approach to interactive image search which enables users to provide feedback via natural language. Based on this work, ~\cite{zhang2019text} proposed a novel constraint augmented reinforcement learning (RL) framework to efficiently incorporate user
preferences over time. These works have several drawbacks: 1) Simply conducting attribute matching is far from understanding users' needs and thus leading to a bad performance. Existence of synonyms or metaphor makes it even harder to retrieve the target image on semantic level. 2) The neural network models employed are not effective. Neural models which gain a lot of research interest such as attention mechanism~\cite{bahdanau2016neural} and  Transformer~\cite{vaswani2017attention} are not leveraged. 3) Fashion attribute information associated with products such as fabric, shape, etc is not fully used in these models. However, these attributes  show great potential in the related tasks such as fashion image modeling, style prediction and so on.

 We propose a framework that can effectively handle conversational fashion image retrieval with multiturn natural language feedback texts mentioned above. 
 One characteristic is that it searches for candidate images based on exploitation of the encoded reference image and feedback text information together with the conversation history via a novel neural framework. It facilitates better predictions based on the information from all previous turns within the concerned session. Another unique component is a comparative analysis module which establishes a relationship between the differential representation derived from the reference image and the candidate image, and the feedback texts contributing to a matching score for the candidate image.
To leverage the fashion attribute information of candidate images, a mutual attention mechanism is designed containing both the attention from the candidate image to the feedback texts and the other way round.
The former attention helps to obtain a flexible feedback text representation according to each candidate image fashion attribute, forming one matching vector for each fashion attribute. The later one aims to adjust the corresponding weights with respect to the matching vectors obtained in the first stage, which contribute to the final matching score.  
Since there is no existing suitable fashion product image dataset that can appropriately capture user search scenario in multiturn settings, we derive a large-scale dataset based on an existing dataset which supports conversational fashion image retrieval with a single-turn natural language feedback. We integrate multiple single-turn data followed by additional manual efforts on a scrutinizing and consistency verification process ensuring that a multiturn session can consistently capture a particular user's fashion product search need.

To sum up, the main contributions of this paper are listed as follows: 
\begin{itemize}
\item Our model interacts with the dialog history from previous turns via a novel neural framework. It also establishes a relationship between the differential representation derived from the reference image and candidate image, and the feedback text contributing to a matching score for the candidate image.
\item Fashion attribute information of candidate images is leveraged via a mutual attention consisting of attention from both candidate image to feedback texts of each turn and the other way round.
\item We derive a new conversational fashion product image retrieval dataset supporting multiturn settings from an existing dataset.
\item Our model outperforms all existing  state-of-the-art methods in the experiment.
\end{itemize}

\begin{figure*}
    \centering
    \includegraphics[height=75mm,width=157mm]{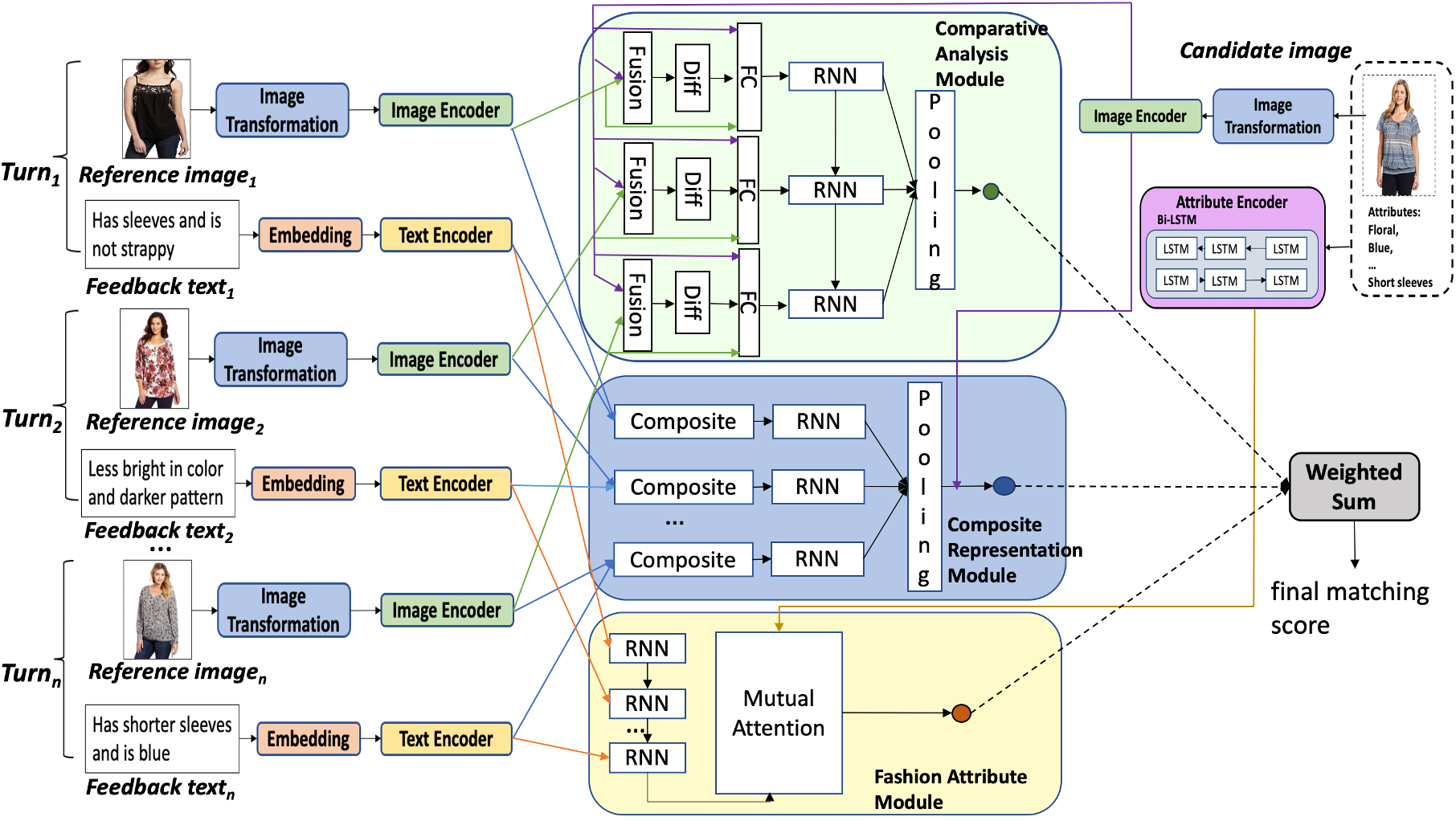}
    \caption{The overall architecture of our framework. It consists of three modules, namely, composite representation module, comparative analysis module, and fashion attribute module. The input of the framework is a conversation context composed of several turns of a session. The output is a matching score corresponding  to the candidate image.}
    \label{fig:overall}
\end{figure*}

\section{Related Work}
\subsection{Conversational Image Retrieval} 
Recent developments in computer vision and natural language processing methods have led to the considerable interest in image retrieval related tasks including image captioning~\cite{rennie2017self,vinyals2015show}, visual question-answering (VQA)~\cite{antol2015vqa,das2018embodied,goyal2017making}, cross-modal image retrieval~\cite{vo2019composing,noh2016image,han2017automatic}. Among these tasks, some research focuses on the topic of image retrieval with natural language feedback. Aiming at selecting the desired image according to natural language feedback, efforts are made by incorporating users' feedback to the reference image and retrieving the image having the highest similarity score. Performance on this task has been enhanced in many  works. For instance, some methods use a predefined set of attribute values to facilitate product retrieval application~\cite{han2017automatic}. Some seek to fuse the image \& text features producing  a more precise representation of the image-text pair ranging from simple techniques (e.g. concatenation, simple feed-forward networks) to advanced techniques such as conducting parameter hashing~\cite{noh2016image}, using a composition classifier~\cite{718510,anwaar2021compositional} or through residual connection~\cite{vo2019composing}. Some methods improve the performance by adding more features such as text-only, image-only, attribute-only  features, etc~\cite{shin2020fashioniq,li2019designovels}. Notably, ~\cite{yu2020curlingnet} gains quite good performance in single-turn conversational image retrieval by adding a correction module which takes the difference between the reference and target image embedding into account.
\subsection{Fashion Search and Recommendation}
When making fashion search decisions, people usually show different preferences for product attributes (e.g. a dress with short sleeves and floral prints), which can correspond to the fashion attributes the target image contain (e.g. sleeves, prints, etc). Fashion recommendation task aims to choose the best-matched product from a large number of fashion products satisfying personalized demands. Along this line, some studies have been proposed to improve the performance of fashion recommendation. In order to model visual characters and user preferences, some methods utilize pre-trained CNN to generate the image representation~\cite{mcauley2015image,he2015vbpr,he2016ups,wu2019hierarchical}. Some methods attempt to get a better understanding of products by leveraging aesthetics and style features~\cite{liu2017deepstyle,yu2018aesthetic}. While some methods manage to provide recommendations following the purpose of explaining the recommendation reason through intuitive fashion attribute semantic highlights in a personalized manner~\cite{hou2019explainable}.

Analyzing fashion attributes is essential in fashion retrieval. Fashion attributes, which include texture, fabric, style, are used to drive the learning of retrieval representations~\cite{al2017fashion,hsiao2017learning,hsiao2018creating}. Typically, fashion attributes extracted from the side information are always informative. In some works, attribute-guided learning is a key factor for retrieval accuracy improvement~\cite{wu2017image,yao2017boosting,you2016image}. Relative attributes, first introduced by ~\cite{parikh2011relative}, can be seen of a supplement of user feedback. Following this concept, a system named 'WhittleSearch' is proposed for fashion image retrieval~\cite{kovashka2012whittlesearch}. When a user states a query, the system calculates the relative strength of attributes to provide the retrieval result. ~\cite{kovashka2013attribute} leverages attributes for guiding relevance feedback information in image search. ~\cite{kovashka2017attributes} aims to discover semantic visual attributes to assist downstream tasks such as image retrieval.

\section{Our Framework}
\subsection{Problem Definition}
As described in the Introduction section, our goal is to find the most relevant image, called the target image, from a collection of candidate images satisfying the user's information need. Typically, each candidate image is associated with some fashion attributes. In each session, the user refines the retrieval result by providing natural language feedback texts. The initial retrieved image is treated as the first reference image, denoted as $p_1$, and the first natural language feedback text is denoted as $t_1$. Given a  dialog context composed of $n$ turns of a session, represented as $(p_1,t_1,...,p_n,t_n)$ where $p_i$ denotes the $i$-th reference image and $t_i$ represents the corresponding feedback texts. In each turn $i$, the feedback texts $t_i$ describes the relative difference between the current reference image $p_i$, and the desired image. The aim is to retrieve the target image $trg$ via computing a matching score with each candidate image in the fashion dataset. 

The training dataset consisting of $N$ sessions is denoted as $D = \{[ref,trg]_i\}_{i=1}^{N}$, where the reference information $ref = (p_1,t_1,...$ $,p_{n_i},t_{n_i})$. The $i$-th session contains $n_i$ turns and each turn is represented as $(p_k,t_k)_{k=1}^{n_i}$ where ${p_k}$ denotes a reference image and ${t_k}$ denotes a user feedback text.

\subsection{Framework Overview}
Figure \ref{fig:overall} depicts the overview of our proposed framework, which consists of three  modules, namely, composite representation module, comparative analysis module, and fashion attribute module. The composite representation module aims to extract and integrate the reference image and feedback text features of each turn and form a composite feature representation. Then it makes an association  between the composite feature with the candidate image representation. Specifically, the image representation is extracted using residual networks. The natural language feedback text of each turn is embedded using  pre-trained word embedding. The embedding of each text is then fed into a Transformer-based self-attention module. By making sentences attend to itself, dependencies of different level are captured. The image and text representation of each turn is composed into one representation and fed into a recurrent network following the turn order with a pooling layer. Finally a partial matching score for the candidate image, which can be treated as a matching score from the perspective of composite features, is obtained.

The comparative analysis module comes up with a differential representation in order to compare the difference between the reference image and the candidate image. The representation is  acquired by feeding the reference image representation, the candidate image representation, as well as the difference of them into a fully connected layer. It then  establishes a relationship between the differential representation and the feedback text by matching the representation with the feedback representation at each turn. After that, the matched vectors of each turn are recorded via a recurrent network with a pooling layer contributing to a partial matching score for the candidate image. 

The fashion attribute module exploits the attribute information of the candidate image and calculates the mutual attention between candidate image and feedback texts. We first embed the fashion attribute texts using the same pre-trained word embedding method as above. Then a recurrent network is applied for the feedback text embedding at each turn. A mutual attention matching method is designed to handle the output of the pooling layer and the candidate image embedding. In the first attention stage, the  feedback texts of all turns are embedded according to different fashion attributes forming one matching vector result for each attribute. In the second attention stage, the corresponding weights are learned with respect to the matching vectors obtained in the first stage. As a result, a partial matching score due to fashion attributes is obtained from the attentive sum of the weighted vectors.

In addition, in order to get a more precise text representation, we employ a self-attention mechanism. By making the textual feedback at each turn attend to itself, intra word-level dependencies are captured. This helps to learn hierarchical representations such as word-level, phrase-level as well as sentence-level representations.

\subsection{Preprocessing and Encoding}
\label{Preprocessing}
\subsubsection{Image Transformation}
Before encoding the images into vectors, we first use some techniques to perform image transformation on product images. Image transformation aims to help make some desired information more obvious or explicit and augment the original data. These techniques include random horizontal flip, random rotation, random translation and random scale, which horizontally flips, rotates, translates, resizes the given image randomly with a given probability. 
\subsubsection{Image Encoder}
\label{image encoder}
After the image transformation, we encode the image representation using ResNet-101 and ResNet-152. We share the parameters for image representation between reference and candidate images. The encoder embeds the $i$-th image from a multiturn session $p_{i}$ into a vector representation $x_{i}^p = ImgEnc(p_{i}) \in  \	\mathbb{R}^{d_p}$. The image encoder is trained from scratch without pretraining on any external data. 
\subsubsection{Spell Checker}
Since the dataset contains some misspelled words (e.g. stripe$\to$strip, colorful$\to$colorfull), we correct the spelling using the {\itshape pyspellchecker}\footnote{https://pypi.org/project/pyspellchecker/} python package if the word cannot be found in the vocabulary. We further manually collect some commonly misspelled fashion-related words to ensure the probability of misspelling is minimized to the lowest. In the experiment, the collection contains 855 fashion-related misspelled words.
\subsubsection{Text Encoder}
\label{text encoder}
To encode the textual feedback of each turn,  the texts are firstly tokenized, and the word embeddings are initialized with GloVe~\cite{pennington2014glove}. For example, the $i$-th feedback of one session $t_i$, after embedded into a vector, is represented as $e_{i}^t\in\mathbb{R}^{l\times{d_t}}$, where $l$ is the number of words in the sentence and $d_t$ is the embedding dimension. The word embeddings are then fed into a self-attention stack encoder. This stack has three inputs: the query sentence $Q\in\mathbb{R}^{l_Q\times{d_t}}$, the key sentence $K\in\mathbb{R}^{l_K\times{d_t}}$, and the value sentence $V\in\mathbb{R}^{l_V\times{d_t}}$. In our case, both of them are equal to the input embedding $e_{i}^t$. 

Specifically, the self-attention encoder is made up of several attention blocks. These blocks have the same structure and are stacked together. Each block takes the output of the former block as input. 
Inside the block, each word in the query sentence is attended to words in the key sentence via Scaled Dot-Product Attention. Then the block multiplies the results to the value input and calculates the weighted sum. Finally it adds the vector to the query sentence and feeds it into fully connected network ~\cite{vaswani2017attention}.

Our feedback text encoding result can be represented as the pooling result of the hierarchical self-attention stack. We use the average pooling, which takes the average of different granularity embedding, as the result.

\begin{equation}
    x_i^t = TxtEnc(t_i) =  Pooling(e_{i,j}^t)_{j=1}^{n_s}\in\mathbb{R}^{d_t}
\end{equation}
\begin{equation}
    e_{i,j}^t = SelfAttention(e_{i,j-1}^t,e_{i,j-1}^t,e_{i,j-1}^t)
\end{equation}
\begin{equation}
    e_{i,0}^t = e_i^t
\end{equation}
where $n_s$ is the number of self-attention blocks, $x_i^t$ represents the encoding result of the feedback text $t_i$, $e_{i,j}$ is the embedding output vector at the $j$-th attention block.
\subsection{Composite Representation Module}
\label{composite representation module}
\subsubsection{Multimodal composer}
At the $i$-th turn, the multimodal composer composes the reference image embedding $x_i^p$ and the feedback text $x_i^t$ into a joint semantic representation $x_i^c$. Here we use ComposeAE
, which is an effective and state-of-the-art method for combining image and text through autoencoding~\cite{anwaar2021compositional}. It takes the image and text embeddings as input, and outputs the composed feature between them. This method shows promising result on image retrieval tasks where texts are used to express the difference between reference and target images. The composition process can be formulated as follows:
\begin{equation}
x_i^c = ComposeAE(x_i^p,x_i^t)
\end{equation}
\subsubsection{Multiturn Analyzer}
\label{Multiturn Analyzer}
The multiturn analyzer is used to aggregate the encoded multimodal representation with the conversation context history from previous turns. In order to better memorize the history information and make decisions based on all turns, a Gated Recurrent Unit ({\bfseries GRU}) network with a pooling layer is employed~\cite{chung2014empirical}. Each time the composed feature is generated, it is then fed into the network following chronological order. We take the output of the pooling layer as the final representation of all multiturn reference image \& text information. The forward formulas of the multiturn analyzer are:
\begin{equation}
    h_i^c = GRU(h_{i-1}^c,x_i^c)
\end{equation}
\begin{equation}
    x^{ref}_{CR} = W_3h_i^c+b_3
\end{equation}
where $h_i^c$ is the hidden representation multimodal feature at the time step $i$, $x^{ref}_{CR}$ is the final representation of all multiturn information, $W_3$ is the parameter we wish to optimize.
\subsubsection{Candidate Generator} 
Given the final representation of all multiturn reference information $x^{ref}_{CR}$, we aim to search the target image from a set of candidate fashion images. Precisely, we calculate the cosine similarity between the reference embedding $x^{ref}_{CR}\in \mathbb{R}^{d_p}$ and the candidate image embedding $x^{cand} \in \mathbb{R}^{d_p}$, which can be obtained from the image encoder described in Section \ref{image encoder}. The cosine similarity then serves as the partial matching score, denoted as $S_1(ref,cand)$, derived from the composition representation module.
\subsection{Comparative Analysis Module}
\subsubsection{Differential Representation}
Aiming at deriving the differential representation between encoded candidate and reference image information, this module first takes the difference between the reference image and candidate image as input. Let $x_i^{a}$ denote the differential representation, the representation is concatenated by three vectors: 
\begin{equation}
    x_i^{diff}= FC([x^{cand}\odot{x_i^p};x^{cand}])-FC([x^{cand}\odot{x_i^p};{x_i^p}])
\end{equation}
\begin{equation}
    x^{a}_i = FC([x_e^{trg};x_e^{ref};x_i^{diff}])
\end{equation}
where $x_i^p$ is the reference image representation at $i$-th turn, $x^{cand}$ is the candidate image representation, $FC$ is a fully connected layer.
\subsubsection{Text Matching}
In order to detect the relationship between the differential representation mentioned above and the feedback text, at each turn, we match the differential representation $x_i^a$ with the feedback text representation $x_i^t$ via element-wise multiplication. The matched vectors of each turn are later fed into a GRU network:
\begin{equation}
    x_i^m = x_i^{a}*{x}_i^t
\end{equation}
\begin{equation}
    h_i^m = GRU(h_{i-1}^m,x_i^m)
\end{equation}
where $x_i^t$ denotes the text encoding at $i$-th turn, $h_i^m$ is the $i$-th hidden layer in the GRU network.

After that, an average pooling layer is employed to record the information of all GRU cells. Finally, the output of the pooling layer is fed into a fully connected layer with RELU activation. The output is served as the partial matching score $S_2(ref,cand)$:
\begin{equation}
    S_2(ref,cand) = g([\bar{h}_i^m]_{i=1}^{n})
\end{equation}
where $[\bar{h}_i^m]_{i=1}^{n}$ is the average pooling output of all hidden layers, and $g(\cdot)$ is a RELU activation function.
\subsection{Fashion Attribute Module}
\subsubsection{Image Fashion Attribute Embedding}
Recall that each candidate image has five  fashion attributes, namely, texture, fabric, shape, part, and style. Each fashion attribute consists of some words describing the corresponding aspect of it.
To encode the candidate images, we also use GloVe (same as in Section \ref{Preprocessing}) to embed each fashion attribute into a $d$ dimension vector. For a candidate image, the embedding is denoted as $(a_{txt},a_{fab},a_{shp},$ $a_{prt},a_{stl})$. Each part of the embedding represents the embedding of one fashion attribute. It is worth noting that some fashion attributes may contain more than one words (e.g. part: long sleeve, button front). We first acquire the embedding of each word, then calculate the average of them as the final fashion attribute embedding.
\subsubsection{Feedback Encoder} 
The feedback text is initiated using the GloVe embedding matrix. The embedding of the feedback text at the $i$-th turn is denoted as $e_i^t$ as mentioned in Section \ref{text encoder}. After that, each feedback is converted into a $d_t$ dimension vector. 

Similar as mentioned in Section \ref{Multiturn Analyzer}, the feedback text information of each turn is recorded chronologically. A bidirectional GRU network is leveraged and takes the embedding of feedback at each turn as input, which can be denoted as:
\begin{equation}
     h_i^{attr}= BiGRU(h_{i-1}^{attr},h_{i+1}^{attr},e_i^{t})
\end{equation}
where $h_{i-1}^{attr}$ is the forward encoded feedback text representation with the history from $i$-1 previous turns, $h_{i+1}^{attr}$ is the backward encoded feedback text representation with the history from $i$+1 previous turns, $e_i^{t}$ is the text representation at $i$-th turn.
\begin{figure}
    \centering
    \includegraphics[width=\linewidth,height=68mm]{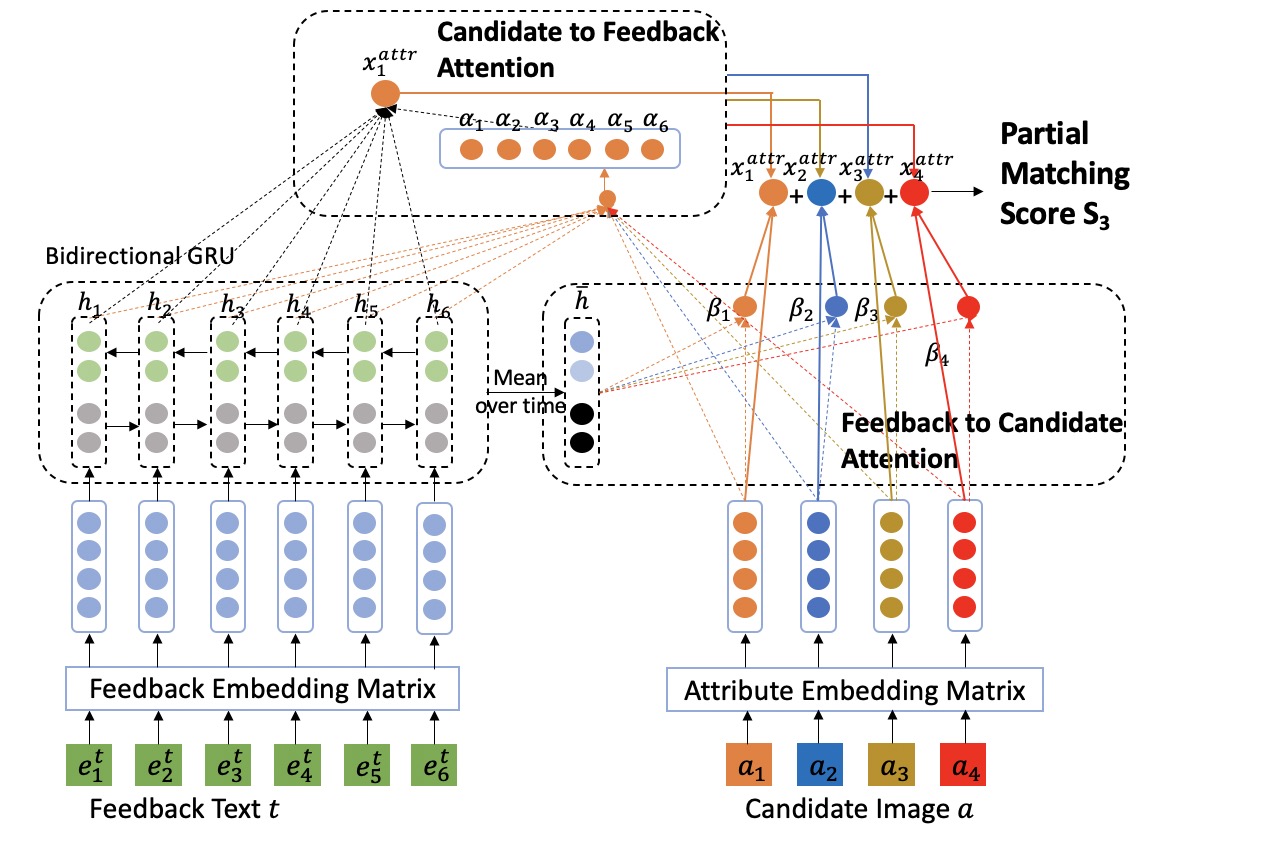}
    \caption{The mutual attention between  the feedback text and the candidate image.}
    \label{fig:my_label}
\end{figure}
\subsubsection{Candidate-to-feedback Attention}
The key element of the fashion attribute module is the mutual attention mechanism. It is composed of mutual attention from the  candidate image to the feedback text and the other way round. 

The candidate-to-feedback attention mainly calculates the attentive embedding of the  feedback text according to each candidate image attribute. When we wish to determine if a candidate image is suitable or not, we first take a look at one of its fashion attributes, fabric for example, then we reread the multiturn feedback to find out which part of the feedback dialog context  should be more focused. Based on this strategy, each candidate fashion attribute should focus on different parts of the  feedback dialog texts. Weights are introduced measuring the relevance between each image fashion attribute and the feedback text at each time step. The following formulas are designed for calculating the weights:
\begin{equation}
    \alpha_{mj} = \frac{exp(w_{mj})}{\sum_{k=1}^{n}exp(w_{mk})}
\end{equation}
\begin{equation}
    w_{mj} = f(W_4^T[a_m;h_j^{attr}]+b_4)
\end{equation}
where $w_{mj}$ denotes the attention weight from the $j$-th turn feedback text hidden layer to the  $m$-th fashion attribute in the candidate image. Note that $a_m\in
\{a_{txt},a_{fab},a_{shp},a_{prt},a_{stl}\}$. $W_4,b_4$ are the parameters to optimize, $f(\cdot)$ is an non-linear activation function. 

According to each image fashion attribute $a_m$, the attention weights are then employed to calculate the weighted sum of the hidden representation in the feedback aspect. The detailed formula is shown as follows:
\begin{equation}
    x_m^{attr}=\sum_{j=1}^n\alpha_{mj}h_j^{attr}
\end{equation}
where $x_m^{attr}$ is the attentive feedback  representation according to the attribute $a_m$.

Having the feedback representation $x_m^{attr}$, we can calculate the similarity between each attentive feedback text representation and candidate attribute representation $a_m$:
\begin{equation}
    S_{attr}(x_m^{attr},a_m) = cosine(x_m^{attr},a_m)
\end{equation}
\subsubsection{Feedback-to-candidate Attention}
Intuitively, for the same feedback information, different values should be put to different image fashion attributes. That is, for the five similarity scores mentioned above, we compute the weight of each one to form the partial similarity score. This leads to the idea of feedback-to-candidate attention, which means that the weights are trained between feedback  representation mentioned in the last section and the candidate attribute representation:
\begin{equation}
    S_3(ref,cand) = \sum\beta_{a_m}S_{attr}(x_m^{attr},a_m)
\end{equation}
\begin{equation}
    \beta_{a_m} = \frac{exp(w_{a_m})}{\sum_{a_k\in\{a_{txt},a_{fab},a_{shp},a_{prt},a_{stl}\}}exp(w_{a_k})}
\end{equation}
\begin{equation}
    w_{a_m} = f(W_5^T[\bar{h}^{attr};a_m]+b_5)
\end{equation}
where $\bar{h}^{attr}$ is the average pooling result of the GRU multiturn analyzer hidden states. $\beta_{e_i}$ is the correlation weight of feedback-to-candidate aspect, indicating which candidate attribute should be more focused in the data.

\subsection{Partial Matching Score Combination}
Given the three partial matching scores $S_1(ref,cand)$, $S_2(ref,cand)$, $S_3(ref,cand)$  mentioned above, the final matching score is computed as the weighted sum of them:
\begin{equation}
    S(ref,cand) = w_1S_1(ref,cand)+w_2S_2(ref,cand)+w_3S_3(ref,cand)
\end{equation}
where $w_1$, $w_2$, $w_3$ are the parameters that need to be optimized.
\subsection{Training}
 During training, we use the max-margin triple loss as the loss function:
\begin{equation}
    L = max(0,margin-S(ref,trg')+S(ref,trg))
\end{equation}
where $margin$ is a hyperparameter which records the gap between positive and negative examples. $trg$ is the true positive target image while $trg'$ is the false negative target image.

It is worth noting that instead of building a negative sample of the target set, we use batch hard triplet loss. For each anchor, we get the hardest positive and negative to form a triplet and build the triplet loss over a batch of embeddings. 

We train three modules separately and record the best score matrix of each module. Then, an iterative process is utilized to ensemble the modules. In the process, the previous best score matrix serves as the new candidate in the next iteration. We use {\itshape hyperopt}\footnote{https://github.com/hyperopt/hyperopt} Bayesian optimization to handle the process, which aims to find optimal weights between partial matching scores and maximize  the overall score.
\section{Experiment}
\subsection{Dataset}
Since there is no existing suitable fashion product dataset which can appropriately capture user modeling in multiturn settings, we derive a large-scale multiturn fashion dataset based on the existing FashionIQ dataset which originally only supports single-turn feedback~\cite{guo2019fashion}. Every fashion product image in the dataset has five fashion  attributes, namely, texture, fabric, shape, part, and style. Each fashion attribute includes  some words describing the corresponding aspect of the image.

The original FashionIQ dataset contains single-turn sessions, and each session is represented by a triplet having the form of (reference image,  feedback text, target image). Many reference and target images in FashionIQ dataset are highly relevant and have duplications. To derive multiturn sessions, we concatenate single-turn sessions in FashionIQ by matching the target image of one triplet with the reference image of another triplet in an automatic manner. For example, the original triplets (img1, txt1, img2), (img2, txt2, img3), (img3, txt3, img4) in FashionIQ dataset can be concatenated into a session having the form of (img1, txt1, img2, txt2, img3, txt3, img4). This process can derive a large number of multiturn sessions. A session with $n$ turns implies that the session has $n-1$ feedback and $n$ images which is composed of $n-1$ reference images and 1 target image. 

However, not all the sessions obtained are reasonable. Therefore, we manually select and filter out the  problematic sessions which belong to several kinds of cases. The first kind refers to duplicates. For example, the current reference image is exactly the same as the next image. The second kind refers to inconsistency. For example, associated with a particular reference image, the feedback is 'white color'. Then a white colored reference image is retrieved. The user continues to write the feedback 'shorter sleeve'. The next reference image retrieved has shorter sleeves than the last one, but is not white colored, which is obviously unreasonable. The third kind refers to conflicts. For example, associated with the first image, the feedback is 'long sleeves'. After the second image is retrieved, the user gives another feedback 'sleeveless'. The last kind is circle. For example, the first reference image is followed by the feedback 'longer sleeves'. After the second reference image is retrieved, the user gives the feedback 'shorter sleeves', the system repeatedly retrieves the reference image same as the first one. 

Moreover, although a fashion attribute dataset is contained in the FashionIQ dataset, the size of the dataset is limited and cannot cover every target image. We expand the fashion attribute dataset to make sure every target image is associated with some fashion attributes following the same way as in ~\cite{guo2019fashion}. Specifically, image fashion attributes were extracted from the product titles, the product summaries, and detailed product descriptions on product websites.
\setlength{\belowcaptionskip}{+0.3cm} 
\begin{table}
  \begin{tabular}{cp{1cm}p{1cm}p{1cm}p{1cm}p{1cm}}
    \toprule
    Category&Sessions with 3 turns&Sessions with 4 turns&Sessions with 5 turns&Total Sessions&Total Images\\
    \midrule
    Dress &3264&763 &311&4338& 5115\\
    Shirt &2821 &687 &167&3675&4130\\
    Toptee &2565 &766 &162&3493&4410\\
    \midrule
    Total & 8650 & 2216 & 640&11506&13655\\
  \bottomrule
\end{tabular}
\caption{Detailed Information of Our Dataset}
  \label{tab:data}
\end{table}
As a result, we gather 11506 multiturn sessions. Table \ref{tab:data} shows the detailed information of our derived dataset. Furthermore, the multiturn sessions are grouped into three categories, namely, dress, toptee, and shirt. Each category includes sessions ranging from 3 to 5 turns. The average length of the feedback text in each turn is 5.02 words.  
\begin{table*}
  \begin{tabular}{cccc|ccc|ccc|ccc}
    \toprule 
    \multirow{2}{*}{Method} & \multicolumn{3}{c}{Overall}&\multicolumn{3}{c}{Dress} &  \multicolumn{3}{c}{Shirt}&\multicolumn{3}{c}{Toptee}\\
     \cline{2-4}\cline{5-7} \cline{8-10}\cline{11-13}
     &R@5& R@8 & MRR & R@5 & R@8 & MRR & R@5&R@8&MRR&R@5&R@8&MRR \\
    \midrule
    Text-only &7.8&12.3&6.9 &7.4&10.8&5.4 &7.5&12.5&6.2&8.2&13.4&6.3 \\
    Image-only &10.7&14.5&6.9&10.2&15.0&6.3&9.3&15.7&5.5&9.1&13.9&7.6\\
    Attribute-only &9.8&12.8&7.7&11.1&13.9&8.2&9.7&11.5&6.8&8.9&11.8&6.3\\
    \midrule
    TIRG ~\cite{vo2019composing}&12.4 &15.9&11.5&12.5&14.2&11.6&13.8&16.7&12.9&12.0&15.6&10.9\\
    RITC ~\cite{shin2020fashioniq}&13.2&18.7&12.4&11.8&18.8&10.2&14.0&20.6&12.2&13.2&18.9&11.7 \\
    ComposeAE ~\cite{anwaar2021compositional}&19.2&25.7&13.4&18.5&26.4&14.4&19.8&25.2&14.5&19.2&26.6&14.8 \\
    CCNet ~\cite{yu2020curlingnet}&13.5&15.5&12.2&12.7&17.2&10.5&15.2&18.5&13.3&13.6&16.2&12.1\\
    AUS ~\cite{guo2019fashion} &11.3&16.2&9.2&13.4&15.3&10.5&14.7&16.6&11.3&12.4&13.3&10.6\\
    Dialog Manager ~\cite{guo2018dialog}& 13.1&15.2&11.6&12.7&16.7&10.8&13.9&17.7&11.6&11.6&15.8&10.3\\
    \midrule
    Ours  &30.3&33.4&26.5&29.8&33.5&25.6&30.5&34.1&27.4&29.4&33.6&26.1\\
    w/o CA &25.5&29.2&21.3&24.1&28.4&18.5&25.6&28.6&22.1&24.5&25.2&21.7\\
    w/o FA &21.3&28.8&17.6&20.5&29.4&15.0&22.9&29.0&18.3&22.0&28.5&17.9\\
    \bottomrule
  \end{tabular}
  \caption{Experimental results on Ours and some comparison methods. The metrics are in percentage. CA denotes the comparative analysis module. FA denotes the fashion attribute module.}
  \label{tab:experiments}
\end{table*}
\subsection{Experimental Setting}
 We prepare 5 runs for the experiments. For each run, we randomly choose 70\% multiturn sessions  from the dataset as the training set and 30\% as the testing set. We conduct our experiments using PyTorch. We use ResNet-152 as our image encoder without pretraining it on any external information. As for the text encoder, we use 3 stacks of self-attention blocks. By default, training is run for 150 epochs with a start learning rate 0.0001. We will release the code and the dataset to the public.

 We use similar evaluation metrics as in previous works ~\cite{guo2019fashion}. Each model outputs K best-matched products having the highest output score for the given session. Since we assume that each session in our dataset corresponds to one true positive image, which is the target image, we calculate the recall rate of the target image among the K selected ones as the main evaluation metric. The metric is denoted as recall at K (R@K), representing  the proportion of target image found in the top-K retrieval results. In our experiment, K is selected to be 5 and 8. We also use MRR (mean reciprocal rank) as another evaluation metric in our experiment. It is the average of the multiplicative inverse of the rank of the target image in all retrieved images.
\subsection{Comparison Models}
In order to evaluate the effectiveness of our proposed model, we compare our model with several baselines and existing state-of-the-art models. Since most of existing models are originally designed for single-turn setting. For fair comparison, we extend these models by adding a recurrent network to aggregate the encoded results with the dialog context from previous turns so that they can handle multiturn settings. 
\paragraph{Text-Only} This model ignores the reference images and retrieves the target image according to user feedback only. The texts are encoded by a LSTM-GRU network and the images are encoded by ResNet-152.
\paragraph{Image-Only} The image-only model only utilizes the image information in the reference image session. It utilizes the information of visual similarity between the information of the visual similarity between candidate and reference images.
\paragraph{Attribute-Only} The attribute-only model treats every image as a combination of fashion attributes. Each fashion attribute is encoded by LSTM-GRU and is concatenated to form the image representation. 
\paragraph{TIRG} A recent method based on concatenation of visual and textual features with an additional gating connection to pass the image features directly to the learned joint feature space ~\cite{vo2019composing}.
\paragraph{RITC} It was first proposed by ~\cite{shin2020fashioniq}. The residual text and image composer is a method that learns the residual between the features of target and  reference images.
\paragraph{ComposeAE} A state-of-the art model proposed by ~\cite{anwaar2021compositional}. It is an autoencoder based model which aims to learn the composition of image and text query
for retrieving images.
\paragraph{Attribute-aware User Simulator (AUS)} It is a model where attribute features are incorporated to augment image representation~\cite{guo2019fashion}.
\paragraph{Correction Network (CCNet)} It is a model that finds the difference between reference and target images and checks its validity with a relative caption~\cite{yu2020curlingnet}. 

\paragraph{Dialog Manager} It was first proposed in ~\cite{guo2018dialog}, which is a framework that considers a user interacting with a retrieval agent via iterative dialog turns. The texts are encoded by a simple LSTM network, and the images are encoded by ResNet. It composes the image and text features by summing them up and feeds them into a memory network. 

\subsection{Experiment Results}
Table \ref{tab:experiments} shows the experiment results of our model as well as all comparison models. Our Model, denoted as Ours, has gained the best R@5, R@8 and MRR score on the whole dataset as well as on all categories.

Generally for most of the models, the R@5, R@8 and MRR rates on the shirt category are higher than the other two categories. One reason is that the appearance of different shirts is less diverse, which assists the models in better capturing the difference between different products. Another reason is that the fashion attribute information of shirts is richer than other two categories. We also conduct the ablation study in our experiment.  Without comparative analysis module, the performance of our model decreases by 4.8\% in R@5, 4.2\% in R@8, and 5.2\% 
in MRR on average. This result reflects the difference between candidate and reference images can be exploited and reflected from the feedback text of each turn. Taken fashion attribute module into account, the performance increases by 9.0\% in R@5, 4.6\% in R@8, 8.9\% in MRR on average. This verifies the effectiveness of fashion attributes for improving retrieval results. Compared with image-only and text-only models, the remaining models utilizing the multimodal feature composer have a better performance (Dialog Manager, RITC, TIRG, ComposeAE, CCNet, AUS), which verifies the necessity of composing the image \& text at each turn into one representation. Compared with models without attention (Text-only, Image-only, Attribute-only, TIRG, ComposeAE, CCNet, AUS, Dialog Manager), attentive models outperform them on the whole, which demonstrates the superior power of attention mechanism in multiturn image retrieval. 
 
\begin{figure}[b]
\centering
\subfigure[Ours]{
\begin{minipage}[t]{0.5\linewidth}
\centering
\includegraphics[scale=0.25]{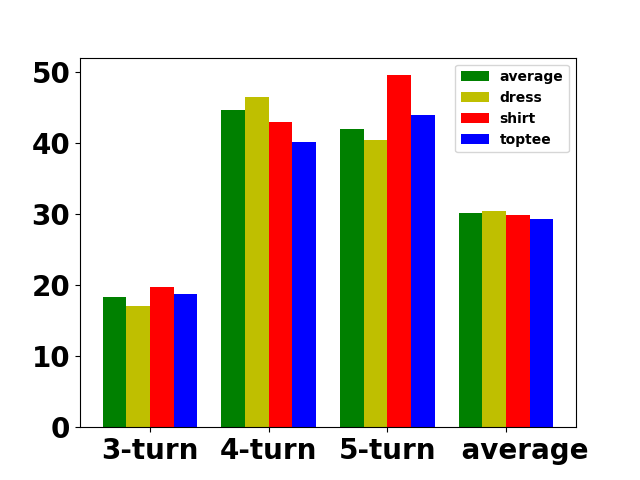}
\label{sub:ours}
\end{minipage}%
}%
\subfigure[ComposeAE]{
\begin{minipage}[t]{0.5\linewidth}
\centering
\includegraphics[scale=0.25]{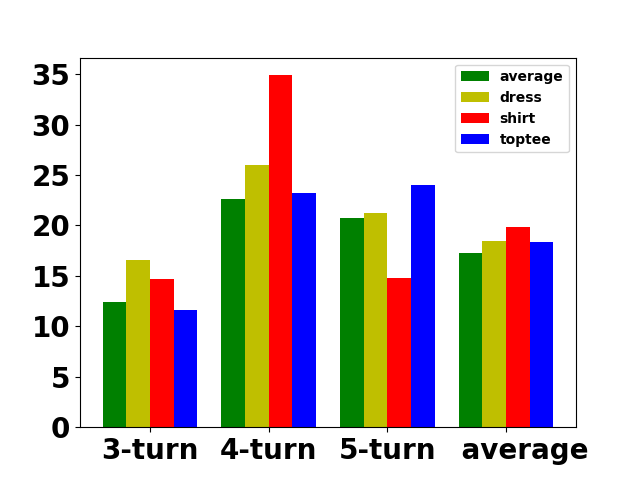}
\label{sub:com}
\end{minipage}%
}%

\subfigure[Attribute-only]{
\begin{minipage}[t]{0.5\linewidth}
\centering
\includegraphics[scale=0.25]{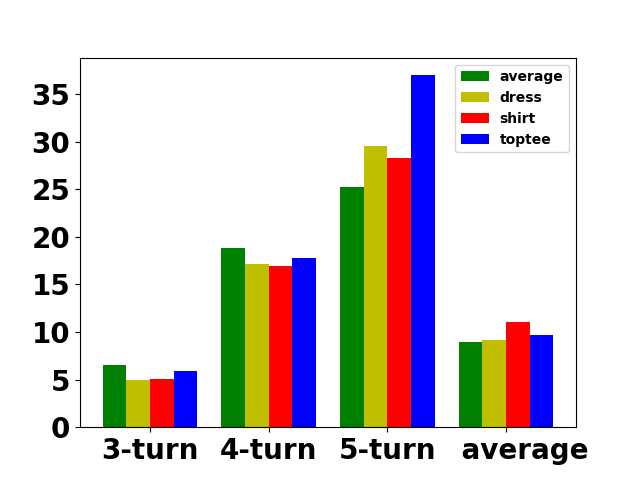}
\label{sub:att}
\end{minipage}
}%
\subfigure[Text-only]{
\begin{minipage}[t]{0.5\linewidth}
\centering
\includegraphics[scale=0.25]{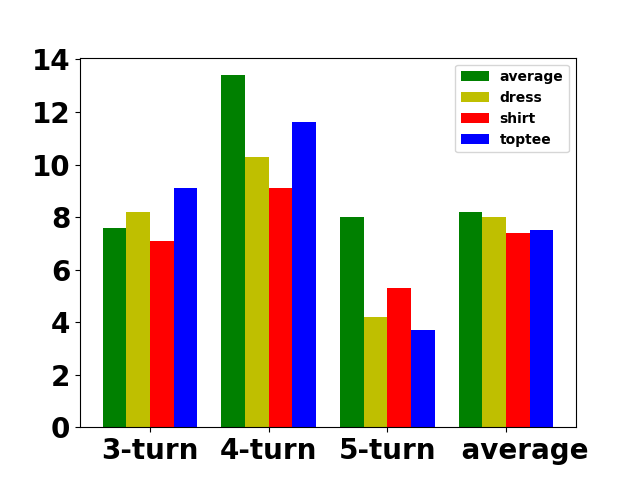}
\label{sub:txt}
\end{minipage}
}%
\centering
\caption{The average R@5 performance  concerning different lengths of turns on different models.}
\label{all}
\end{figure}
\subsection{Further Analysis} 
 We study how our model performs under different turn lengths. We also conduct similar investigation for three baselines. Recall that our dataset contains sessions ranging from 3 to 5 turns, Figure \ref{all} shows the analysis on R@5 rate with respect to different turn lengths.
 
For attribute-only model, according to Figure \ref{sub:att}, the longer the session is, the better the performance will be. The reason is that the fashion attribute information is insufficient in short sessions. While for text-only models, according to \ref{sub:txt},  the performance on 5-turn sessions is worse compared to the other two 
types. For the reason that on one hand, the number of 5-turn length sessions is much smaller than the other two types. On the other hand, without image information for reference, it is rather hard for the model to understand the target image the user wishes to retrieve, especially when it comes to long conversations.
 \begin{figure*}[t]
    \centering
    \includegraphics[width=170mm,height=45mm]{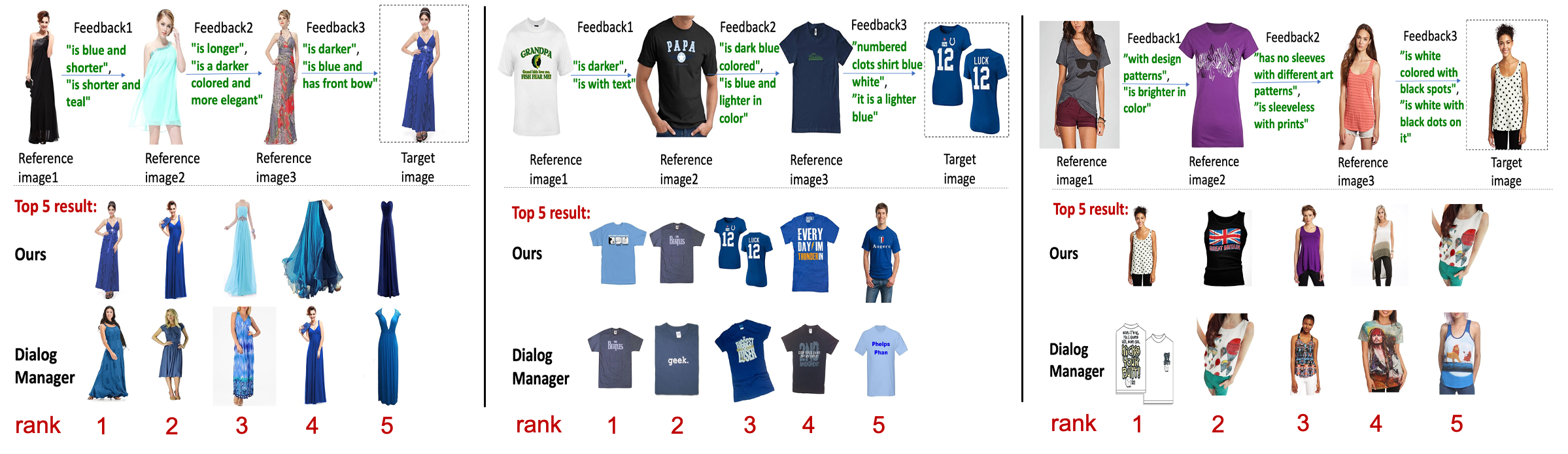}
    \caption{Some case studies.}
    \label{fig:case study}
\end{figure*}
According to Figure \ref{sub:com}, on all categories, the 4-turn sessions have the best performance in the  ComposeAE model, while the 3-turn sessions have the worst. However, in our model, the performances on 4 and 5-turn sessions are similar. The  4 and 5-turn sessions have improved performance in the two models because  they have richer information. Through interacting with the system, more hints are provided in the user feedback, which implies that if the session is too short, the information provided in each turn may not be rich enough for accurately retrieving the target image. However, for the ComposeAE model, if the session is too long, it may give rise to the problem of information lost. Information of previous turns is easier to be forgotten in the former time steps. This problem gets tackled by the fashion attribute module in our model where the mutual attention between candidate image and feedback captures the information of each turn according to image attributes. The information buried in long sessions can be fully utilized by the model. 
 


\subsection{Case Study} We collect some cases to analyze our retrieval results compared with a comparison model, namely Dialog Manager as shown in Figure \ref{fig:case study}. Five products are retrieved by our model and Dialog Manager, respectively. It can be observed that the retrieved image by our model can satisfy the user's need, which shows the effectiveness of the proposed model. Specifically, the desired target images in the first and third case are  ranked the first among all candidate images,  while they do not appear in the top-5 list in the comparison model result. Particularly, in the second case, the target image is ranked the third in our model while is ranked the ninth in the comparison model.
Interestingly, general requirements such as 'blue colored','lighter' can be recognized by both models. However, detailed requirements such as 'has front bow', 'numbered slots', 'black spots' are challenging. One reason for the efficacy of our model is that it exploits the utilization of the fashion attribute module. Detailed information such as 'bow', 'number print', 'dots' can often be found clues in image fashion attributes, thus facilitating the interaction between target image and user feedback text. Another reason is the adoption of the self-attention feedback encoder. By making feedback text attend to itself, richer information of different granularities is learned.  

Compared with Dialog Manager, our model is also  better at capturing the information from previous turns. In the first case, the feedback 'is shorter' seems to be forgotten in the comparison model result. The retrieved images do not have the 'shorter' aspect compared with the first target image. In the third case, 'sleeveless' is a very important clue in retrieving target images. However, not all the images retrieved by Dialog Manager top-5 list are sleeveless. One of the reasons that our model has better performance is that in our model, the fashion attribute module containing mutual attention is employed to connect the information from previous turns to the present task. Furthermore, due to the comparative analysis module, the retrieval results of our model is more similar to the reference images. In cases where there are many candidate images fulfilling the feedback requirements, such as the second case, choosing the one which is more  similar to the reference image will produce the results with higher quality. 

\section{Conclusions} In our paper, we investigate multiturn conversational fashion image retrieval with natural language feedback. A new framework is proposed which can effectively handle the task. Our model searches for target images based on the aggregation of the encoded reference image and text information  with the conversation history via a novel neural framework. Moreover, it utilizes fashion attribute to improve the performance. We also derive a dataset suitable for multiturn conversational fashion image retrieval. Empirical results demonstrate the effectiveness of our model. The results show that our approach outperforms all the baselines and state-of-the-art models. In our future work, we wish to expand the generality of the approach on other fashion items having more wholistic fashion attributes and explore the usability of the work in real-world fashion domain applications.



\bibliographystyle{ACM-Reference-Format}
\balance
\bibliography{sample-sigconf}


\begin{thebibliography}{46}


\ifx \showCODEN    \undefined \def \showCODEN     #1{\unskip}     \fi
\ifx \showDOI      \undefined \def \showDOI       #1{#1}\fi
\ifx \showISBNx    \undefined \def \showISBNx     #1{\unskip}     \fi
\ifx \showISBNxiii \undefined \def \showISBNxiii  #1{\unskip}     \fi
\ifx \showISSN     \undefined \def \showISSN      #1{\unskip}     \fi
\ifx \showLCCN     \undefined \def \showLCCN      #1{\unskip}     \fi
\ifx \shownote     \undefined \def \shownote      #1{#1}          \fi
\ifx \showarticletitle \undefined \def \showarticletitle #1{#1}   \fi
\ifx \showURL      \undefined \def \showURL       {\relax}        \fi
\providecommand\bibfield[2]{#2}
\providecommand\bibinfo[2]{#2}
\providecommand\natexlab[1]{#1}
\providecommand\showeprint[2][]{arXiv:#2}

\bibitem[\protect\citeauthoryear{Al-Halah, Stiefelhagen, and Grauman}{Al-Halah
  et~al\mbox{.}}{2017}]%
        {al2017fashion}
\bibfield{author}{\bibinfo{person}{Ziad Al-Halah}, \bibinfo{person}{Rainer
  Stiefelhagen}, {and} \bibinfo{person}{Kristen Grauman}.}
  \bibinfo{year}{2017}\natexlab{}.
\newblock \showarticletitle{Fashion Forward: Forecasting Visual Style in
  Fashion}. In \bibinfo{booktitle}{\emph{Proceedings of the IEEE international
  conference on computer vision}}. \bibinfo{pages}{388--397}.
\newblock


\bibitem[\protect\citeauthoryear{Antol, Agrawal, Lu, Mitchell, Batra,
  Lawrence~Zitnick, and Parikh}{Antol et~al\mbox{.}}{2015}]%
        {antol2015vqa}
\bibfield{author}{\bibinfo{person}{Stanislaw Antol}, \bibinfo{person}{Aishwarya
  Agrawal}, \bibinfo{person}{Jiasen Lu}, \bibinfo{person}{Margaret Mitchell},
  \bibinfo{person}{Dhruv Batra}, \bibinfo{person}{C Lawrence~Zitnick}, {and}
  \bibinfo{person}{Devi Parikh}.} \bibinfo{year}{2015}\natexlab{}.
\newblock \showarticletitle{Vqa: Visual Question Answering}. In
  \bibinfo{booktitle}{\emph{Proceedings of the IEEE international conference on
  computer vision}}. \bibinfo{pages}{2425--2433}.
\newblock


\bibitem[\protect\citeauthoryear{Anwaar, Labintcev, and Kleinsteuber}{Anwaar
  et~al\mbox{.}}{2021}]%
        {anwaar2021compositional}
\bibfield{author}{\bibinfo{person}{Muhammad~Umer Anwaar}, \bibinfo{person}{Egor
  Labintcev}, {and} \bibinfo{person}{Martin Kleinsteuber}.}
  \bibinfo{year}{2021}\natexlab{}.
\newblock \showarticletitle{Compositional Learning of Image-Text Query for
  Image Retrieval}. In \bibinfo{booktitle}{\emph{Proceedings of the IEEE/CVF
  Winter Conference on Applications of Computer Vision}}.
  \bibinfo{pages}{1140--1149}.
\newblock


\bibitem[\protect\citeauthoryear{Bahdanau, Cho, and Bengio}{Bahdanau
  et~al\mbox{.}}{2016}]%
        {bahdanau2016neural}
\bibfield{author}{\bibinfo{person}{Dzmitry Bahdanau},
  \bibinfo{person}{Kyunghyun Cho}, {and} \bibinfo{person}{Yoshua Bengio}.}
  \bibinfo{year}{2016}\natexlab{}.
\newblock \bibinfo{title}{Neural Machine Translation by Jointly Learning to
  Align and Translate}.
\newblock
\newblock
\showeprint[arxiv]{1409.0473}~[cs.CL]


\bibitem[\protect\citeauthoryear{Chung, Gulcehre, Cho, and Bengio}{Chung
  et~al\mbox{.}}{2014}]%
        {chung2014empirical}
\bibfield{author}{\bibinfo{person}{Junyoung Chung}, \bibinfo{person}{Caglar
  Gulcehre}, \bibinfo{person}{KyungHyun Cho}, {and} \bibinfo{person}{Yoshua
  Bengio}.} \bibinfo{year}{2014}\natexlab{}.
\newblock \showarticletitle{Empirical Evaluation of Gated Recurrent Neural
  Networks on Sequence Modeling}.
\newblock \bibinfo{journal}{\emph{arXiv Preprint arXiv:1412.3555}}
  (\bibinfo{year}{2014}).
\newblock


\bibitem[\protect\citeauthoryear{Das, Datta, Gkioxari, Lee, Parikh, and
  Batra}{Das et~al\mbox{.}}{2018}]%
        {das2018embodied}
\bibfield{author}{\bibinfo{person}{Abhishek Das}, \bibinfo{person}{Samyak
  Datta}, \bibinfo{person}{Georgia Gkioxari}, \bibinfo{person}{Stefan Lee},
  \bibinfo{person}{Devi Parikh}, {and} \bibinfo{person}{Dhruv Batra}.}
  \bibinfo{year}{2018}\natexlab{}.
\newblock \showarticletitle{Embodied Question Answering}. In
  \bibinfo{booktitle}{\emph{Proceedings of the IEEE Conference on Computer
  Vision and Pattern Recognition Workshops}}. \bibinfo{pages}{2054--2063}.
\newblock


\bibitem[\protect\citeauthoryear{Datta, Joshi, Li, and Wang}{Datta
  et~al\mbox{.}}{2008}]%
        {datta2008image}
\bibfield{author}{\bibinfo{person}{Ritendra Datta}, \bibinfo{person}{Dhiraj
  Joshi}, \bibinfo{person}{Jia Li}, {and} \bibinfo{person}{James~Z Wang}.}
  \bibinfo{year}{2008}\natexlab{}.
\newblock \showarticletitle{Image Retrieval: Ideas, Influences, and Trends of
  the New Age}.
\newblock \bibinfo{journal}{\emph{Comput. Surveys}} \bibinfo{volume}{40},
  \bibinfo{number}{2} (\bibinfo{year}{2008}), \bibinfo{pages}{1--60}.
\newblock


\bibitem[\protect\citeauthoryear{Goyal, Khot, Summers-Stay, Batra, and
  Parikh}{Goyal et~al\mbox{.}}{2017}]%
        {goyal2017making}
\bibfield{author}{\bibinfo{person}{Yash Goyal}, \bibinfo{person}{Tejas Khot},
  \bibinfo{person}{Douglas Summers-Stay}, \bibinfo{person}{Dhruv Batra}, {and}
  \bibinfo{person}{Devi Parikh}.} \bibinfo{year}{2017}\natexlab{}.
\newblock \showarticletitle{Making the v in VQA Matter: Elevating the Role of
  Image Understanding in Visual Question Answering}. In
  \bibinfo{booktitle}{\emph{Proceedings of the IEEE Conference on Computer
  Vision and Pattern Recognition}}. \bibinfo{pages}{6904--6913}.
\newblock


\bibitem[\protect\citeauthoryear{Guo, Wu, Cheng, Rennie, Tesauro, and
  Feris}{Guo et~al\mbox{.}}{2018}]%
        {guo2018dialog}
\bibfield{author}{\bibinfo{person}{Xiaoxiao Guo}, \bibinfo{person}{Hui Wu},
  \bibinfo{person}{Yu Cheng}, \bibinfo{person}{Steven Rennie},
  \bibinfo{person}{Gerald Tesauro}, {and} \bibinfo{person}{Rogerio Feris}.}
  \bibinfo{year}{2018}\natexlab{}.
\newblock \showarticletitle{Dialog-Based Interactive Image Retrieval}. In
  \bibinfo{booktitle}{\emph{Advances in neural information processing
  systems}}. \bibinfo{pages}{678--688}.
\newblock


\bibitem[\protect\citeauthoryear{Guo, Wu, Gao, Rennie, and Feris}{Guo
  et~al\mbox{.}}{2019}]%
        {guo2019fashion}
\bibfield{author}{\bibinfo{person}{Xiaoxiao Guo}, \bibinfo{person}{Hui Wu},
  \bibinfo{person}{Yupeng Gao}, \bibinfo{person}{Steven Rennie}, {and}
  \bibinfo{person}{Rogerio Feris}.} \bibinfo{year}{2019}\natexlab{}.
\newblock \showarticletitle{Fashion IQ: A New Dataset Towards Retrieving Images
  by Natural Language Feedback}.
\newblock \bibinfo{journal}{\emph{arXiv Preprint arXiv:1905.12794}}
  (\bibinfo{year}{2019}).
\newblock


\bibitem[\protect\citeauthoryear{Han, Wu, Huang, Zhang, Zhu, Li, Zhao, and
  Davis}{Han et~al\mbox{.}}{2017a}]%
        {han2017automatic}
\bibfield{author}{\bibinfo{person}{Xintong Han}, \bibinfo{person}{Zuxuan Wu},
  \bibinfo{person}{Phoenix~X Huang}, \bibinfo{person}{Xiao Zhang},
  \bibinfo{person}{Menglong Zhu}, \bibinfo{person}{Yuan Li},
  \bibinfo{person}{Yang Zhao}, {and} \bibinfo{person}{Larry~S Davis}.}
  \bibinfo{year}{2017}\natexlab{a}.
\newblock \showarticletitle{Automatic Spatially-Aware Fashion Concept
  Discovery}. In \bibinfo{booktitle}{\emph{Proceedings of the IEEE
  International Conference on Computer Vision}}. \bibinfo{pages}{1463--1471}.
\newblock


\bibitem[\protect\citeauthoryear{Han, Wu, Jiang, and Davis}{Han
  et~al\mbox{.}}{2017b}]%
        {han2017learning}
\bibfield{author}{\bibinfo{person}{Xintong Han}, \bibinfo{person}{Zuxuan Wu},
  \bibinfo{person}{Yu-Gang Jiang}, {and} \bibinfo{person}{Larry~S Davis}.}
  \bibinfo{year}{2017}\natexlab{b}.
\newblock \showarticletitle{Learning Fashion Compatibility With Bidirectional
  LSTMS}. In \bibinfo{booktitle}{\emph{Proceedings of the 25th ACM
  international conference on Multimedia}}. \bibinfo{pages}{1078--1086}.
\newblock


\bibitem[\protect\citeauthoryear{He and McAuley}{He and McAuley}{2015}]%
        {he2015vbpr}
\bibfield{author}{\bibinfo{person}{Ruining He} {and} \bibinfo{person}{Julian
  McAuley}.} \bibinfo{year}{2015}\natexlab{}.
\newblock \showarticletitle{VBPR: Visual Bayesian Personalized Ranking From
  Implicit Feedback}.
\newblock \bibinfo{journal}{\emph{arXiv Preprint arXiv:1510.01784}}
  (\bibinfo{year}{2015}).
\newblock


\bibitem[\protect\citeauthoryear{He and McAuley}{He and McAuley}{2016}]%
        {he2016ups}
\bibfield{author}{\bibinfo{person}{Ruining He} {and} \bibinfo{person}{Julian
  McAuley}.} \bibinfo{year}{2016}\natexlab{}.
\newblock \showarticletitle{Ups and Downs: Modeling the Visual Evolution of
  Fashion Trends With One-Class Collaborative Filtering}. In
  \bibinfo{booktitle}{\emph{proceedings of the 25th international conference on
  world wide web}}. \bibinfo{pages}{507--517}.
\newblock


\bibitem[\protect\citeauthoryear{Hou, Wu, Chen, Li, Zheng, and Liu}{Hou
  et~al\mbox{.}}{2019}]%
        {hou2019explainable}
\bibfield{author}{\bibinfo{person}{Min Hou}, \bibinfo{person}{Le Wu},
  \bibinfo{person}{Enhong Chen}, \bibinfo{person}{Zhi Li},
  \bibinfo{person}{Vincent~W. Zheng}, {and} \bibinfo{person}{Qi Liu}.}
  \bibinfo{year}{2019}\natexlab{}.
\newblock \bibinfo{title}{Explainable Fashion Recommendation: A Semantic
  Attribute Region Guided Approach}.
\newblock
\newblock
\showeprint[arxiv]{1905.12862}~[cs.IR]


\bibitem[\protect\citeauthoryear{Hsiao and Grauman}{Hsiao and Grauman}{2017}]%
        {hsiao2017learning}
\bibfield{author}{\bibinfo{person}{Wei-Lin Hsiao} {and}
  \bibinfo{person}{Kristen Grauman}.} \bibinfo{year}{2017}\natexlab{}.
\newblock \showarticletitle{Learning the Latent “Look”: Unsupervised
  Discovery of a Style-Coherent Embedding From Fashion Images}. In
  \bibinfo{booktitle}{\emph{2017 IEEE International Conference on Computer
  Vision (ICCV)}}. IEEE, \bibinfo{pages}{4213--4222}.
\newblock


\bibitem[\protect\citeauthoryear{Hsiao and Grauman}{Hsiao and Grauman}{2018}]%
        {hsiao2018creating}
\bibfield{author}{\bibinfo{person}{Wei-Lin Hsiao} {and}
  \bibinfo{person}{Kristen Grauman}.} \bibinfo{year}{2018}\natexlab{}.
\newblock \showarticletitle{Creating Capsule Wardrobes From Fashion Images}. In
  \bibinfo{booktitle}{\emph{Proceedings of the IEEE Conference on Computer
  Vision and Pattern Recognition}}. \bibinfo{pages}{7161--7170}.
\newblock


\bibitem[\protect\citeauthoryear{Hu, Yi, and Davis}{Hu et~al\mbox{.}}{2015}]%
        {hu2015collaborative}
\bibfield{author}{\bibinfo{person}{Yang Hu}, \bibinfo{person}{Xi Yi}, {and}
  \bibinfo{person}{Larry~S Davis}.} \bibinfo{year}{2015}\natexlab{}.
\newblock \showarticletitle{Collaborative Fashion Recommendation: A Functional
  Tensor Factorization Approach}. In \bibinfo{booktitle}{\emph{Proceedings of
  the 23rd ACM international conference on Multimedia}}.
  \bibinfo{pages}{129--138}.
\newblock


\bibitem[\protect\citeauthoryear{Kovashka and Grauman}{Kovashka and
  Grauman}{2013}]%
        {kovashka2013attribute}
\bibfield{author}{\bibinfo{person}{Adriana Kovashka} {and}
  \bibinfo{person}{Kristen Grauman}.} \bibinfo{year}{2013}\natexlab{}.
\newblock \showarticletitle{Attribute Pivots for Guiding Relevance Feedback in
  Image Search}. In \bibinfo{booktitle}{\emph{Proceedings of the IEEE
  International Conference on Computer Vision}}. \bibinfo{pages}{297--304}.
\newblock


\bibitem[\protect\citeauthoryear{Kovashka and Grauman}{Kovashka and
  Grauman}{2017}]%
        {kovashka2017attributes}
\bibfield{author}{\bibinfo{person}{Adriana Kovashka} {and}
  \bibinfo{person}{Kristen Grauman}.} \bibinfo{year}{2017}\natexlab{}.
\newblock \showarticletitle{Attributes for Image Retrieval}.
\newblock In \bibinfo{booktitle}{\emph{Visual Attributes}}.
  \bibinfo{publisher}{Springer}, \bibinfo{pages}{89--117}.
\newblock


\bibitem[\protect\citeauthoryear{Kovashka, Parikh, and Grauman}{Kovashka
  et~al\mbox{.}}{2012}]%
        {kovashka2012whittlesearch}
\bibfield{author}{\bibinfo{person}{Adriana Kovashka}, \bibinfo{person}{Devi
  Parikh}, {and} \bibinfo{person}{Kristen Grauman}.}
  \bibinfo{year}{2012}\natexlab{}.
\newblock \showarticletitle{Whittlesearch: Image Search With Relative Attribute
  Feedback}. In \bibinfo{booktitle}{\emph{2012 IEEE Conference on Computer
  Vision and Pattern Recognition}}. IEEE, \bibinfo{pages}{2973--2980}.
\newblock


\bibitem[\protect\citeauthoryear{Li, whan Lee, sang Song, young Shin, and hyun
  Go}{Li et~al\mbox{.}}{2019}]%
        {li2019designovels}
\bibfield{author}{\bibinfo{person}{Jianri Li}, \bibinfo{person}{Jae whan Lee},
  \bibinfo{person}{Woo sang Song}, \bibinfo{person}{Ki young Shin}, {and}
  \bibinfo{person}{Byung hyun Go}.} \bibinfo{year}{2019}\natexlab{}.
\newblock \bibinfo{title}{Designovel's System Description for Fashion-Iq
  Challenge 2019}.
\newblock
\newblock
\showeprint[arxiv]{1910.11119}~[cs.CV]


\bibitem[\protect\citeauthoryear{Li, Cao, Zhu, and Luo}{Li
  et~al\mbox{.}}{2017}]%
        {li2017mining}
\bibfield{author}{\bibinfo{person}{Yuncheng Li}, \bibinfo{person}{Liangliang
  Cao}, \bibinfo{person}{Jiang Zhu}, {and} \bibinfo{person}{Jiebo Luo}.}
  \bibinfo{year}{2017}\natexlab{}.
\newblock \showarticletitle{Mining Fashion Outfit Composition Using an
  End-to-End Deep Learning Approach on Set Data}.
\newblock \bibinfo{journal}{\emph{IEEE Transactions on Multimedia}}
  \bibinfo{volume}{19}, \bibinfo{number}{8} (\bibinfo{year}{2017}),
  \bibinfo{pages}{1946--1955}.
\newblock


\bibitem[\protect\citeauthoryear{Liao, Ma, He, Hong, and Chua}{Liao
  et~al\mbox{.}}{2018}]%
        {liao2018knowledge}
\bibfield{author}{\bibinfo{person}{Lizi Liao}, \bibinfo{person}{Yunshan Ma},
  \bibinfo{person}{Xiangnan He}, \bibinfo{person}{Richang Hong}, {and}
  \bibinfo{person}{Tat-seng Chua}.} \bibinfo{year}{2018}\natexlab{}.
\newblock \showarticletitle{Knowledge-Aware Multimodal Dialogue Systems}. In
  \bibinfo{booktitle}{\emph{Proceedings of the 26th ACM international
  conference on Multimedia}}. \bibinfo{pages}{801--809}.
\newblock


\bibitem[\protect\citeauthoryear{Liu, Wu, and Wang}{Liu et~al\mbox{.}}{2017}]%
        {liu2017deepstyle}
\bibfield{author}{\bibinfo{person}{Qiang Liu}, \bibinfo{person}{Shu Wu}, {and}
  \bibinfo{person}{Liang Wang}.} \bibinfo{year}{2017}\natexlab{}.
\newblock \showarticletitle{DeepStyle: Learning User Preferences for Visual
  Recommendation}. In \bibinfo{booktitle}{\emph{Proceedings of the 40th
  International ACM SIGIR Conference on Research and Development in Information
  Retrieval}}. \bibinfo{pages}{841--844}.
\newblock


\bibitem[\protect\citeauthoryear{Liu, Feng, Song, Zhang, Lu, Xu, and Yan}{Liu
  et~al\mbox{.}}{2012}]%
        {liu2012hi}
\bibfield{author}{\bibinfo{person}{Si Liu}, \bibinfo{person}{Jiashi Feng},
  \bibinfo{person}{Zheng Song}, \bibinfo{person}{Tianzhu Zhang},
  \bibinfo{person}{Hanqing Lu}, \bibinfo{person}{Changsheng Xu}, {and}
  \bibinfo{person}{Shuicheng Yan}.} \bibinfo{year}{2012}\natexlab{}.
\newblock \showarticletitle{Hi, Magic Closet, Tell Me What to Wear!}. In
  \bibinfo{booktitle}{\emph{Proceedings of the 20th ACM international
  conference on Multimedia}}. \bibinfo{pages}{619--628}.
\newblock


\bibitem[\protect\citeauthoryear{Ma, Jia, Zhou, Fu, Liu, and Tong}{Ma
  et~al\mbox{.}}{2017}]%
        {ma2017towards}
\bibfield{author}{\bibinfo{person}{Yihui Ma}, \bibinfo{person}{Jia Jia},
  \bibinfo{person}{Suping Zhou}, \bibinfo{person}{Jingtian Fu},
  \bibinfo{person}{Yejun Liu}, {and} \bibinfo{person}{Zijian Tong}.}
  \bibinfo{year}{2017}\natexlab{}.
\newblock \showarticletitle{Towards Better Understanding the Clothing Fashion
  Styles: A Multimodal Deep Learning Approach}. In
  \bibinfo{booktitle}{\emph{Thirty-First AAAI Conference on Artificial
  Intelligence}}.
\newblock


\bibitem[\protect\citeauthoryear{McAuley, Targett, Shi, and Van
  Den~Hengel}{McAuley et~al\mbox{.}}{2015}]%
        {mcauley2015image}
\bibfield{author}{\bibinfo{person}{Julian McAuley},
  \bibinfo{person}{Christopher Targett}, \bibinfo{person}{Qinfeng Shi}, {and}
  \bibinfo{person}{Anton Van Den~Hengel}.} \bibinfo{year}{2015}\natexlab{}.
\newblock \showarticletitle{Image-Based Recommendations on Styles and
  Substitutes}. In \bibinfo{booktitle}{\emph{Proceedings of the 38th
  international ACM SIGIR conference on research and development in information
  retrieval}}. \bibinfo{pages}{43--52}.
\newblock


\bibitem[\protect\citeauthoryear{Noh, Hongsuck~Seo, and Han}{Noh
  et~al\mbox{.}}{2016}]%
        {noh2016image}
\bibfield{author}{\bibinfo{person}{Hyeonwoo Noh}, \bibinfo{person}{Paul
  Hongsuck~Seo}, {and} \bibinfo{person}{Bohyung Han}.}
  \bibinfo{year}{2016}\natexlab{}.
\newblock \showarticletitle{Image Question Answering Using Convolutional Neural
  Network With Dynamic Parameter Prediction}. In
  \bibinfo{booktitle}{\emph{Proceedings of the IEEE conference on computer
  vision and pattern recognition}}. \bibinfo{pages}{30--38}.
\newblock


\bibitem[\protect\citeauthoryear{Parikh and Grauman}{Parikh and
  Grauman}{2011}]%
        {parikh2011relative}
\bibfield{author}{\bibinfo{person}{Devi Parikh} {and} \bibinfo{person}{Kristen
  Grauman}.} \bibinfo{year}{2011}\natexlab{}.
\newblock \showarticletitle{Relative Attributes}. In
  \bibinfo{booktitle}{\emph{2011 International Conference on Computer Vision}}.
  IEEE, \bibinfo{pages}{503--510}.
\newblock


\bibitem[\protect\citeauthoryear{Pennington, Socher, and Manning}{Pennington
  et~al\mbox{.}}{2014}]%
        {pennington2014glove}
\bibfield{author}{\bibinfo{person}{Jeffrey Pennington},
  \bibinfo{person}{Richard Socher}, {and} \bibinfo{person}{Christopher~D
  Manning}.} \bibinfo{year}{2014}\natexlab{}.
\newblock \showarticletitle{Glove: Global Vectors for Word Representation}. In
  \bibinfo{booktitle}{\emph{Proceedings of the 2014 conference on empirical
  methods in natural language processing (EMNLP)}}.
  \bibinfo{pages}{1532--1543}.
\newblock


\bibitem[\protect\citeauthoryear{Rennie, Marcheret, Mroueh, Ross, and
  Goel}{Rennie et~al\mbox{.}}{2017}]%
        {rennie2017self}
\bibfield{author}{\bibinfo{person}{Steven~J Rennie}, \bibinfo{person}{Etienne
  Marcheret}, \bibinfo{person}{Youssef Mroueh}, \bibinfo{person}{Jerret Ross},
  {and} \bibinfo{person}{Vaibhava Goel}.} \bibinfo{year}{2017}\natexlab{}.
\newblock \showarticletitle{Self-Critical Sequence Training for Image
  Captioning}. In \bibinfo{booktitle}{\emph{Proceedings of the IEEE Conference
  on Computer Vision and Pattern Recognition}}. \bibinfo{pages}{7008--7024}.
\newblock


\bibitem[\protect\citeauthoryear{Rostamzadeh, Hosseini, Boquet, Stokowiec,
  Zhang, Jauvin, and Pal}{Rostamzadeh et~al\mbox{.}}{2018}]%
        {rostamzadeh2018fashion}
\bibfield{author}{\bibinfo{person}{Negar Rostamzadeh},
  \bibinfo{person}{Seyedarian Hosseini}, \bibinfo{person}{Thomas Boquet},
  \bibinfo{person}{Wojciech Stokowiec}, \bibinfo{person}{Ying Zhang},
  \bibinfo{person}{Christian Jauvin}, {and} \bibinfo{person}{Chris Pal}.}
  \bibinfo{year}{2018}\natexlab{}.
\newblock \showarticletitle{Fashion-Gen: The Generative Fashion Dataset and
  Challenge}.
\newblock \bibinfo{journal}{\emph{arXiv Preprint arXiv:1806.08317}}
  (\bibinfo{year}{2018}).
\newblock


\bibitem[\protect\citeauthoryear{Shin, Cho, and Hong}{Shin
  et~al\mbox{.}}{2020}]%
        {shin2020fashioniq}
\bibfield{author}{\bibinfo{person}{Minchul Shin}, \bibinfo{person}{Yoonjae
  Cho}, {and} \bibinfo{person}{Seongwuk Hong}.}
  \bibinfo{year}{2020}\natexlab{}.
\newblock \bibinfo{title}{Fashion-Iq 2020 Challenge 2nd Place Team's Solution}.
\newblock
\newblock
\showeprint[arxiv]{2007.06404}~[cs.CV]


\bibitem[\protect\citeauthoryear{Vaswani, Shazeer, Parmar, Uszkoreit, Jones,
  Gomez, Kaiser, and Polosukhin}{Vaswani et~al\mbox{.}}{2017}]%
        {vaswani2017attention}
\bibfield{author}{\bibinfo{person}{Ashish Vaswani}, \bibinfo{person}{Noam
  Shazeer}, \bibinfo{person}{Niki Parmar}, \bibinfo{person}{Jakob Uszkoreit},
  \bibinfo{person}{Llion Jones}, \bibinfo{person}{Aidan~N Gomez},
  \bibinfo{person}{{\L}ukasz Kaiser}, {and} \bibinfo{person}{Illia
  Polosukhin}.} \bibinfo{year}{2017}\natexlab{}.
\newblock \showarticletitle{Attention Is All You Need}. In
  \bibinfo{booktitle}{\emph{Advances in neural information processing
  systems}}. \bibinfo{pages}{5998--6008}.
\newblock


\bibitem[\protect\citeauthoryear{Vinyals, Toshev, Bengio, and Erhan}{Vinyals
  et~al\mbox{.}}{2015}]%
        {vinyals2015show}
\bibfield{author}{\bibinfo{person}{Oriol Vinyals}, \bibinfo{person}{Alexander
  Toshev}, \bibinfo{person}{Samy Bengio}, {and} \bibinfo{person}{Dumitru
  Erhan}.} \bibinfo{year}{2015}\natexlab{}.
\newblock \showarticletitle{Show and Tell: A Neural Image Caption Generator}.
  In \bibinfo{booktitle}{\emph{Proceedings of the IEEE conference on computer
  vision and pattern recognition}}. \bibinfo{pages}{3156--3164}.
\newblock


\bibitem[\protect\citeauthoryear{Vo, Jiang, Sun, Murphy, Li, Fei-Fei, and
  Hays}{Vo et~al\mbox{.}}{2019}]%
        {vo2019composing}
\bibfield{author}{\bibinfo{person}{Nam Vo}, \bibinfo{person}{Lu Jiang},
  \bibinfo{person}{Chen Sun}, \bibinfo{person}{Kevin Murphy},
  \bibinfo{person}{Li-Jia Li}, \bibinfo{person}{Li Fei-Fei}, {and}
  \bibinfo{person}{James Hays}.} \bibinfo{year}{2019}\natexlab{}.
\newblock \showarticletitle{Composing Text and Image for Image Retrieval-an
  Empirical Odyssey}. In \bibinfo{booktitle}{\emph{Proceedings of the IEEE
  Conference on Computer Vision and Pattern Recognition}}.
  \bibinfo{pages}{6439--6448}.
\newblock


\bibitem[\protect\citeauthoryear{Wu, Chen, Hong, Fu, Xie, and Wang}{Wu
  et~al\mbox{.}}{2019}]%
        {wu2019hierarchical}
\bibfield{author}{\bibinfo{person}{Le Wu}, \bibinfo{person}{Lei Chen},
  \bibinfo{person}{Richang Hong}, \bibinfo{person}{Yanjie Fu},
  \bibinfo{person}{Xing Xie}, {and} \bibinfo{person}{Meng Wang}.}
  \bibinfo{year}{2019}\natexlab{}.
\newblock \showarticletitle{A Hierarchical Attention Model for Social
  Contextual Image Recommendation}.
\newblock \bibinfo{journal}{\emph{IEEE Transactions on Knowledge and Data
  Engineering}} (\bibinfo{year}{2019}).
\newblock


\bibitem[\protect\citeauthoryear{Wu, Shen, Wang, Dick, and van~den Hengel}{Wu
  et~al\mbox{.}}{2017}]%
        {wu2017image}
\bibfield{author}{\bibinfo{person}{Qi Wu}, \bibinfo{person}{Chunhua Shen},
  \bibinfo{person}{Peng Wang}, \bibinfo{person}{Anthony Dick}, {and}
  \bibinfo{person}{Anton van~den Hengel}.} \bibinfo{year}{2017}\natexlab{}.
\newblock \showarticletitle{Image Captioning and Visual Question Answering
  Based on Attributes and External Knowledge}.
\newblock \bibinfo{journal}{\emph{IEEE Transactions on Pattern Analysis and
  Machine Intelligence}} \bibinfo{volume}{40}, \bibinfo{number}{6}
  (\bibinfo{year}{2017}), \bibinfo{pages}{1367--1381}.
\newblock


\bibitem[\protect\citeauthoryear{Yao, Pan, Li, Qiu, and Mei}{Yao
  et~al\mbox{.}}{2017}]%
        {yao2017boosting}
\bibfield{author}{\bibinfo{person}{Ting Yao}, \bibinfo{person}{Yingwei Pan},
  \bibinfo{person}{Yehao Li}, \bibinfo{person}{Zhaofan Qiu}, {and}
  \bibinfo{person}{Tao Mei}.} \bibinfo{year}{2017}\natexlab{}.
\newblock \showarticletitle{Boosting Image Captioning With Attributes}. In
  \bibinfo{booktitle}{\emph{Proceedings of the IEEE International Conference on
  Computer Vision}}. \bibinfo{pages}{4894--4902}.
\newblock


\bibitem[\protect\citeauthoryear{{Yong Rui}, {Huang}, {Ortega}, and
  {Mehrotra}}{{Yong Rui} et~al\mbox{.}}{1998}]%
        {718510}
\bibfield{author}{\bibinfo{person}{{Yong Rui}}, \bibinfo{person}{T.~S.
  {Huang}}, \bibinfo{person}{M. {Ortega}}, {and} \bibinfo{person}{S.
  {Mehrotra}}.} \bibinfo{year}{1998}\natexlab{}.
\newblock \showarticletitle{Relevance Feedback: A Power Tool for Interactive
  Content-Based Image Retrieval}.
\newblock \bibinfo{journal}{\emph{IEEE Transactions on Circuits and Systems for
  Video Technology}} \bibinfo{volume}{8}, \bibinfo{number}{5}
  (\bibinfo{year}{1998}), \bibinfo{pages}{644--655}.
\newblock


\bibitem[\protect\citeauthoryear{You, Jin, Wang, Fang, and Luo}{You
  et~al\mbox{.}}{2016}]%
        {you2016image}
\bibfield{author}{\bibinfo{person}{Quanzeng You}, \bibinfo{person}{Hailin Jin},
  \bibinfo{person}{Zhaowen Wang}, \bibinfo{person}{Chen Fang}, {and}
  \bibinfo{person}{Jiebo Luo}.} \bibinfo{year}{2016}\natexlab{}.
\newblock \showarticletitle{Image Captioning With Semantic Attention}. In
  \bibinfo{booktitle}{\emph{Proceedings of the IEEE conference on computer
  vision and pattern recognition}}. \bibinfo{pages}{4651--4659}.
\newblock


\bibitem[\protect\citeauthoryear{Yu, Zhang, He, Chen, Xiong, and Qin}{Yu
  et~al\mbox{.}}{2018}]%
        {yu2018aesthetic}
\bibfield{author}{\bibinfo{person}{Wenhui Yu}, \bibinfo{person}{Huidi Zhang},
  \bibinfo{person}{Xiangnan He}, \bibinfo{person}{Xu Chen}, \bibinfo{person}{Li
  Xiong}, {and} \bibinfo{person}{Zheng Qin}.} \bibinfo{year}{2018}\natexlab{}.
\newblock \showarticletitle{Aesthetic-Based Clothing Recommendation}. In
  \bibinfo{booktitle}{\emph{Proceedings of the 2018 World Wide Web
  Conference}}. \bibinfo{pages}{649--658}.
\newblock


\bibitem[\protect\citeauthoryear{Yu, Lee, Choi, and Kim}{Yu
  et~al\mbox{.}}{2020}]%
        {yu2020curlingnet}
\bibfield{author}{\bibinfo{person}{Youngjae Yu}, \bibinfo{person}{Seunghwan
  Lee}, \bibinfo{person}{Yuncheol Choi}, {and} \bibinfo{person}{Gunhee Kim}.}
  \bibinfo{year}{2020}\natexlab{}.
\newblock \bibinfo{title}{CurlingNet: Compositional Learning Between Images and
  Text for Fashion IQ Data}.
\newblock
\newblock
\showeprint[arxiv]{2003.12299}~[cs.CV]


\bibitem[\protect\citeauthoryear{Zhang, Yu, Shen, Jin, Chen, and Carin}{Zhang
  et~al\mbox{.}}{2019}]%
        {zhang2019text}
\bibfield{author}{\bibinfo{person}{Ruiyi Zhang}, \bibinfo{person}{Tong Yu},
  \bibinfo{person}{Yilin Shen}, \bibinfo{person}{Hongxia Jin},
  \bibinfo{person}{Changyou Chen}, {and} \bibinfo{person}{Lawrence Carin}.}
  \bibinfo{year}{2019}\natexlab{}.
\newblock \showarticletitle{Text-Based Interactive Recommendation With
  Constraint-Augmented Reinforcement Learning}.
\newblock  (\bibinfo{year}{2019}).
\newblock


\bibitem[\protect\citeauthoryear{Zhang, Yu, Shen, Jin, Chen, and Carin}{Zhang
  et~al\mbox{.}}{2020}]%
        {zhang2020reward}
\bibfield{author}{\bibinfo{person}{Ruiyi Zhang}, \bibinfo{person}{Tong Yu},
  \bibinfo{person}{Yilin Shen}, \bibinfo{person}{Hongxia Jin},
  \bibinfo{person}{Changyou Chen}, {and} \bibinfo{person}{Lawrence Carin}.}
  \bibinfo{year}{2020}\natexlab{}.
\newblock \showarticletitle{Reward Constrained Interactive Recommendation With
  Natural Language Feedback}.
\newblock \bibinfo{journal}{\emph{arXiv Preprint arXiv:2005.01618}}
  (\bibinfo{year}{2020}).
\newblock


\end{thebibliography}


\end{document}